\def\year{2022}\relax
\documentclass[letterpaper]{article} 
\usepackage{aaai22}  
\usepackage{times}  
\usepackage{helvet}  
\usepackage{courier}  
\usepackage[hyphens]{url}  
\usepackage{graphicx} 
\usepackage{natbib}  
\usepackage{caption} 
\DeclareCaptionStyle{ruled}{labelfont=normalfont,labelsep=colon,strut=off} 
\usepackage{subfigure}
\usepackage{algorithm}
\usepackage[noend]{algpseudocode}
\usepackage{amsmath}
\newtheorem{property}{Property}
\newtheorem{definition}{Definition}
\usepackage{multirow}
\usepackage{newfloat}
\usepackage{listings}
\floatstyle{ruled}
\newfloat{listing}{tb}{lst}{}
\floatname{listing}{Listing}
\nocopyright
\title{Computing Class Hierarchies from Classifiers}
\author{
Kai Kang and Fangzhen Lin
}
\affiliations{

    Department of Computer Science and Engineering\\
    Hong Kong University of Science and Technology\\
Hong Kong\\
kkangab@connect.ust.hk, flin@cs.ust.hk
}

\begin{document}

\maketitle

\begin{abstract}
A class or taxonomic hierarchy is often manually constructed, and part of our knowledge
about the world. In this paper, we propose a novel algorithm for automatically acquiring
a class hierarchy from a classifier which is
often a large neural network these days. The information that we need from a
classifier is its confusion matrix which contains, for each pair of base classes,
the number of errors the classifier makes by mistaking one for another.
Our algorithm produces surprisingly good hierarchies for
some well-known
deep neural network models trained on the CIFAR-10 dataset,
a neural network model for predicting the native language of a
non-native English speaker, a neural network model for detecting the language of a written text,
and a classifier for identifying music genre. In the literature, such class hierarchies
have been used to provide interpretability to the neural networks. We also discuss some other
potential uses of the acquired hierarchies.
\end{abstract}

\section{Introduction}
Much of our knowledge about the world is organized hierarchically. We have classes or
concepts like chairs, tables, office furniture, plastic furniture, furniture, and
so on. Similarly, we may classify animals into cats, dogs, mountain lions, small dogs,
pets, and so on. Many, if not most, of these concepts are not precisely defined, and
are classified by features like their functionality and their visual appearance to us.
With the advance of deep learning, we now have computer
programs that can recognise many of these classes with an accuracy that 
rivals human perception. However, these programs
are often flat and end-to-end, in the sense that they just classify the input into
a class without referencing to any possible hierarchies among
these classes.

There has been work on leveraging our knowledge about class hierarchies to
improve the performance of deep neural network classification models in terms of
accuracies, and sampling and training efficiencies \cite{brust2019integrating, marszalek2007semantic, gao2011discriminative, li2019large, pan2020scoring}.
In this paper, our goal is somewhat the opposite. We take a good
classification model as a given black box, and try to derive from it
a plausible class hierarchy network. Our basic assumption is that the classification
model is accurate enough so that the ``closeness'' or ``relatedness'' of two
classes are directly related to how often the model misclassifies one of them as the other.

We can see several potential uses of a class hierarchy constructed from a classification
model. Firstly, the derived class hierarchy can be used as a sort of justification
or explanation of the given classification model, as some previous work \cite{frosst2017distilling, wan2020nbdt} has
argued. Secondly, a good class hierarchy is useful in its own right. For example, given
a good model to identify the native language of a non-native English speaker, we can
construct a hierarchy of languages that can help to group non-native English speakers
for the purpose of teaching English as a foreign language. We will give several examples
of this nature when discussing experiments of this work.

As we mentioned earlier,
our knowledge about the world is often organized hierarchically. Thus coming up with a good
class hierarchy can be considered a part of scientific discovery. Hopefully our proposed algorithm
for acquiring such knowledge from an accurate
neural network classification model can contribute to this process of
scientific discovery.

This rest of this paper is organized as follows. In the next section, we 
describe the formal setup of the problem and give our method for solving it.
We then show two intuitive properties of our algorithm. Next we apply our algorithm to several
domains, and discuss our expeperimental results.
We then discuss related work and conclude the paper.

\section{Our Method}
\begin{figure}[h]
\subfigure[]
{
	\begin{minipage}[t]{0.22\textwidth}
	\centering        
	\includegraphics[height=1in]{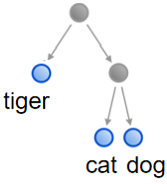}  
	\label{fig:treedemoa}
	\end{minipage}
}
\subfigure[]
{
	\begin{minipage}[t]{0.22\textwidth}
	\centering     
	\includegraphics[height=1in]{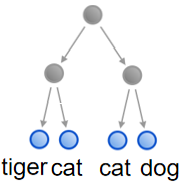}
	\label{fig:treedemob}
	\end{minipage}
}


\caption{(a) A single inheritance tree; (b) A multiple inheritance tree}
\label{fig:treedemo}
\end{figure}

Throughout this paper, we use numerical numbers starting at $0$ to denote base classes.
Given a set $C$ of $n$ base classes $\{0,...,{n-1}\}$, we consider ways to organize them
into
a hierarchy. In graph theoretic terms,  a class hierarchy is a labeled tree where
the label of a node is the class that the node represents, and
the parent to children edges stand for subsumption. So if $N$ is a child
node of $N'$, then the class represented by node $N$
is a subclass of that represented by $N'$. 
Thus for a tree to be a class hierarchy for $C$, its leaves must be labeled by
base classes in $C$, and its internal nodes by classes not in $C$. While every class
in $C$ must be a label of a leave, it can be the labels of multiple leaves.
If a class tree has no distinct nodes labeled
by the same class, then there is no overlapping between any of its subclasses.
In this case, we call the class tree a Single Inheritance Tree (SIT)
as there is a unique path from the root class to any of its subclasses. Otherwise, it
is called a Multiple Inheritance Tree (MIT). 
The following example illustrates the difference between these two types of
hierarchies.
Consider $C=\{0,1,2\}$ with $0$ for cat, $1$ for dog, and $2$ for tiger.
The class $cat$ is related to both $dog$ and $tiger$.
A reasonable way to take account of this is to create a super class for $cat$ and $dog$,
and another for $cat$ and $tiger$. This will yield a MIT as shown in Figure~\ref{fig:treedemob}.
On the other hand, if we insist on using a SIT, and assume that $cat$ is closer to $dog$,
we may come up with the SIT as shown in Figure~\ref{fig:treedemoa}. 

We propose an iterative algorithm to construct a hierarchy for a given set
of base classes by first merging some base classes into super-classes, and then
iteratively merging the remaining base classes and the new classes into more general classes until a single most general class is obtained. The general idea is that classes are merged into a new class if and only if they have similar properties. The question is of course how to define similarity. For this, we will use a similarity matrix that measures the degree of similarities between any two classes, and an adjustable threshold to define whether the degrees of similarity between two classes are large enough to be merged into a new class.

To compute the degree of similarities between two classes, we assume a classification model for the base classes. It can be a decision tree, a neuron network, a set of rules or others. The only thing we need from the model is its so called confusion matrix \cite{Fawcett06}: the confusion matrix ${\mathbf M}$ of a model for $n$ classes is a $n\times n$ integer matrix where for any $0\leq i,j\leq n-1$, the element ${\mathbf M}_{ij}$ is the number of samples in the testing set that are labeled as class $i$ but classified as class $j$ by the model. Often, the sample set is not balanced, so we use a normalized version of the confusion matrix. Specifically, if ${\mathbf M}$ is a confusion matrix, then its normalized version, written $\widehat{{\mathbf M}}$, is defined as: for each $0 \leq i,j\leq n-1$,

\[ \widehat{{\mathbf M}}_{ij} = \frac{{\mathbf M}_{ij} }{\sum^{n-1}_{k = 0} {\mathbf M}_{ik}}.
\]
Notice that $\sum^{n-1}_{k = 0} {\mathbf M}_{ik}$ is the total number of class $i$ instances used to test the classifier to derive the confusion matrix. So $\widehat{{\mathbf M}}_{ij}$ is in fact the error rate of the model for confusing class $i$ instances as class $j$'s, and our normalized confusion matrix can be called the confusion rate matrix.

A confusion matrix is often not symmetric in the sense that for many $i$ and $j$, ${\mathbf M}_{ij} \neq {\mathbf M}_{ji}$, meaning the number of class $i$ instances that are mis-classified as class $j$ is not equal to
the number of class $j$ instances that are mis-classified as class $i$.
However, for the purpose of class hierarchy, we want a number to measure the similarity of
two classes, and this number should be symmetric. This leads to our following definition of
similarity matrices.

\begin{definition}
The similarity matrix ${\mathbf S}$ of a classifier is
\[
\mathbf{S} = \frac{1}{2}(\widehat{{\mathbf M}}+\widehat{{\mathbf M}}^T),\]
where $\widehat{{\mathbf M}}$ is the normalized confusion matrix of the classifier.
\end{definition}

Intuitively, ${\mathbf S}_{ij}=\frac{1}{2}(\widehat{{\mathbf M}}_{ij}+\widehat{{\mathbf M}}_{ji})$ is
the average confusion rate between classes $i$ and $j$, and
the higher it is, the more similar the classes $i$ and $j$ are, and
more likely they are related and inherited from a common ancestor. But it is
domain dependent that just how
large the value ${\mathbf S}_{ij}$ should be to trigger the decision that they are likely siblings
and thus should be merged to form a superclass. To account for this, we introduce a
threshold ratio $r$, and merge classes $i$ and $j$ if the following condition holds:
\begin{equation}
  {\mathbf S}_{ij} > 0 \textbf{ and } {\mathbf S}_{ij} + \delta \geq \max(\{{\mathbf S}_{i,k}| k \neq i\} \cup \{{\mathbf S}_{j,k}| k \neq j\})
\label{eq:merge}
\end{equation}
where $\delta =  r * m = r * \max \{{\mathbf S}_{i,j}|0\leq i<j\leq n-1\}$. It is possible that a class may need to be merged with two other classes. In this case, three of them will be
merged. This means that our computed class hierarchy may not be a binary tree.

The condition (\ref{eq:merge}) will apply not only to base classes but also to newly created
superclasses, with a more involved definition for $m$ to compute the $\delta$.
This implies that we will need a way to compute the similarity scores between
pairs of superclasses created during the process of constructing a hierarchy.
This can be computed from the scores between the superclasses' subclasses by
taking the min, the max, the average, or using some other
formulas. 
Intuitively, taking the min 
is a conservative approach, the max aggressive, and the average in the middle.
However, 
by choosing an appropriate threshold ratio $r$ mentioned earlier, we can twist
the middle approach to make it either more conservative or more aggressive.
So in this paper when computing the similarity between two classes, we take
the average of the similarity scores between their subclasses.
Indeed, as we shall see, it works well for the domains that we have tried so far.

Figure \ref{alg:CH} outlines our algorithm for computing a class hierarchy given the
similarity matrix of a classifier. 
The main function is $CH({\mathbf {S}},r,flag)$, which takes ${\mathbf S}$, the similarity matrix of the classifier defined earlier, $r\geq 0$, the threshold ratio $r$ mentioned above,
and $flag$, a boolean number to indicate whether the returned class hierarchy must be of single inheritance. 
It initializes $\widehat{\mathbf S}$ to ${\mathbf S}$, and $H$ to $\{ t_0,...,t_{n-1}\}$, where $t_i$ is the singleton tree for base class $i$.
It then loops through $MERGE(H,\widehat{\mathbf S},\delta,flag)$, which merges some classes
 in $H$ to create some new superclasses, and
$SIMILARITY(H,\widehat{\mathbf S})$, which computes a new similarity matrix for the new set $H$
of classes. Notice that we identify a tree in $H$ with the class denoted by its root.
The loop exits when $MERGE()$ does not produce any new class, either because the current
$H$ contains just one tree or none of the trees in $H$ can be merged. In the former case, the tree in $H$ is then returned as the computed hierarchy. In the latter case, all trees in $H$ are combined
to form a single tree.

Notice that $m$ for computing $\delta$ in $CH()$ is initialized first for base classes.
It is then updated (line 8) by taking the maximum of similarity scores other than those with
overlapping subtrees. Nodes with overlapping subtrees can indeed be generated when
computing MITs. The similarity scores between these nodes will be close to large, often
close to 1 
like ${\mathbf S}_{ii}$, so should be excluded when computing the threshold $\delta$.

Both $SIMILARITY()$ and $MERGE()$ functions are sketched in Figure~\ref{alg:CH}. Crucially,
$FIND\_ALL\_PAIRS()$ returns a list of
all pairs of class indices $(i,j)$ that satisfy condition (\ref{eq:merge}),
and this list is passed to 
$PAIRS\_TO\_GRAPHS()$ to return a partition of $H$, with each set in the partition representing
the set of classes that are to be merged into a superclass.
The complete code of our algorithm is given in an appendix uploaded as supplementary material.

\begin{figure}[htbp]
\centering
\includegraphics[width=0.3\textwidth]{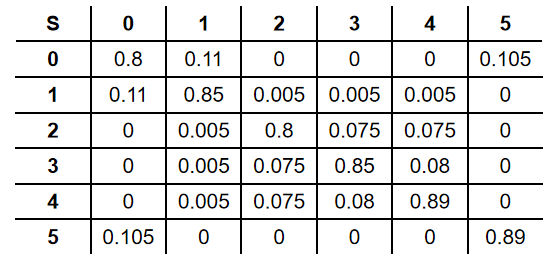} 
\caption{A similarity matrix demo.}
\label{fig:matrixexample}
\end{figure}

To illustrate, consider a case with 6 base classes and a
confusion matrix that produces the initial
similarity matrix given in Figure \ref{fig:matrixexample}.
Initially $H=\{ t_0,..,t_5\}$. $MERGE()$ uses condition (\ref{eq:merge}) to decide which
trees (classes) in $H$ are to be merged. 
Suppose we let $r=0.1$. Then
\[
m =  \max(\{{\mathbf S}_{i,j}|0\leq i<j\leq n-1\}) = 0.11
\] 
\[
\delta =  r * m = 0.011
\] 
So by (\ref{eq:merge}), the pairs in the following list
could be merged:
\[
[(t_0, t_1), (t_0, t_5), (t_3,t_4), (t_2, t_3), (t_2, t_4)].
\]
We see that some of these pairs have common elements so a decision needs to be made whether
to further merge them. In this case, our algorithm will merge
$(t_3,t_4), (t_2, t_3), (t_2, t_4)$ into $(t_2,t_3,t_4)$, thus creating a new tree with
three children trees $t_2$, $t_3$ and $t_4$. However, it won't merge
$(t_0,t_1)$ and $(t_0,t_5)$ into $(t_0,t_1, t_5)$ because the pair $(t_1, t_5)$ is not in the
above list as classes $1$ and $5$ are not similar enough.

If SIT is desired, the above pair list will be converted into $[(t_0, t_1), (t_2, t_3, t_4)]$.
Notice that while both $(t_0, t_5)$ and $(t_0, t_1)$ are in the above list,
only the latter is selected as ${\mathbf S_{01}} > {\mathbf S_{05}}$, meaning class
$0$ is more similar to class $1$ than class $5$.

\begin{figure}[h]
\centering  
\subfigure[]
{
	\begin{minipage}[t]{0.22\textwidth}
	\centering        
	\includegraphics[width=1.5in]{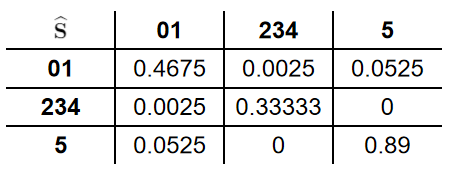}  
	\label{fig:sitdemoS}
	\end{minipage}
}
\subfigure[]
{
	\begin{minipage}[t]{0.22\textwidth}
	\centering     
	\includegraphics[width=1.5in]{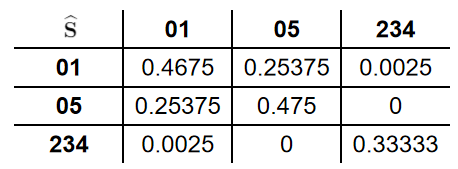}
	\label{fig:mitdemoS}
	\end{minipage}
}
\caption{Updated similarity matrix: (a) SIT, and (b) MIT}
\label{fig:demoS}
\end{figure}

\begin{figure}[h]
\subfigure[]
{
	\begin{minipage}[t]{0.22\textwidth}
	\centering        
	\includegraphics[height=1in]{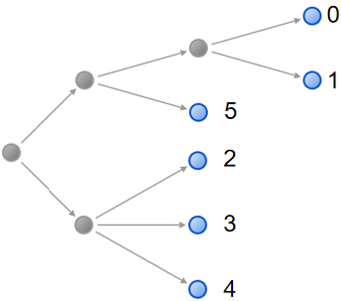}  
	\label{fig:sitdemo}
	\end{minipage}
}
\subfigure[]
{
	\begin{minipage}[t]{0.22\textwidth}
	\centering     
	\includegraphics[height=1in]{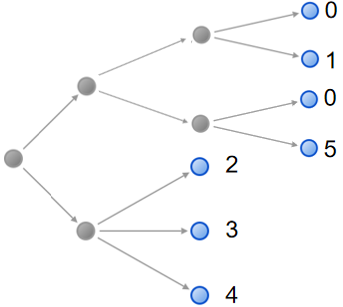}
	\label{fig:mitdemo}
	\end{minipage}
}
\caption{Hierarchies for $\mathbf S$ in Figure \ref{fig:matrixexample}: (a) SIT; (b) MIT}
\label{fig:treedemo2}
\end{figure}

 So $H$ will be updated to be the following set of trees $\{t_{01}, t_{234}, t_5\}$, and its
 similarity matrix computed as given in Figure \ref{fig:sitdemoS}. Next step is to combine $(t_{01}, t_5)$ to get $t_{01\_5}$, and then combine it with $t_{234}$ to return the tree as shown in Figure~\ref{fig:sitdemo}. 

However, if MIT is required, the pair list will be converted into $[(t_0, t_1), (t_0, t_5), (t_2, t_3, t_4)]$, both of two pairs related to $t_0$ will be saved because overlapping is allowed. Then the $H$ will be updated to $\{t_{01}, t_{05}, t_{234}\}$. The new similarity matrix will be figure \ref{fig:mitdemoS}. At next iteration, $(t_{01}, t_{05})$ will be combined and get a new superclass $t_{01\_05}$ then it will be combined with $t_{234}$ to obtain the tree as shown in Figure~\ref{fig:mitdemo}.

\begin{figure}[tp]
\vspace*{-\baselineskip}

\begin{minipage}{\columnwidth}
\begin{algorithm}[H]
\begin{algorithmic}[1] 
\Require 
${\mathbf S}$ -  similarity matrix;
$r$ - threshold ratio, a non-negative real for calculating clustering sensitivity;
$flag$ - 1 if the output must be of single inheritance, 0 otherwise.
\Ensure 
a class hierarchy for $C$.
\Function{CH}{${\mathbf S},r,flag$} (main function)
\State $H = \{ t_0,...,t_{n-1}\}$
\State ${\widehat{\mathbf S}} = {\mathbf S}$
\State $m = \max(\{\widehat{{\mathbf S}}_{i,j}|0\leq i<j\leq n-1\})$
\While{$len(H) > 1$ \textbf{and} $m > 0$}
\State $H = \Call{merge}{H,\widehat{{\mathbf S}}, r * m, flag}$
\State $\widehat{{\mathbf S}} = \Call{similarity}{H, \widehat{{\mathbf S}}}$
\State $m = \max\{\widehat{{\mathbf S}}_{i,j}$ for all $0 \leq i < j < len(H)$ such that $H[i]$ and $H[j]$ has no overlap$\}$ or $0$
\EndWhile
\If {$len(H) > 1$}
\State $H =$ Combine all trees in $H$
\EndIf
\State \Return{$H$} 
\EndFunction
\end{algorithmic} 
\end{algorithm}

\begin{algorithm}[H]
\begin{algorithmic}[1] 
\Require 
Tree set: $H$, similarity matrix: $\widehat{{\mathbf S}}$, threshold: $\delta$, tree type flag: $flag$.
\Ensure 
Updated tree set $H$.
\Function{Merge}{$H, \widehat{{\mathbf S}}, \delta, flag$}
\State $P =  \Call{Find\_all\_pairs}{\widehat{{\mathbf S}}, \delta}$
\State $G =  \Call{pairs\_to\_graphs}{\widehat{{\mathbf S}}, P, flag}$ 
\State $H' \gets []$
\State $rest \gets [0] * len(H)$
\For{$graph$ \textbf{in} $G$} 
\State $tree \gets$ Combine trees indicated by $graph$.
\State $H'.append(tree)$ 
\For{$i$ \textbf{in} $graph$}
\State $rest[i] = 1$
\EndFor
\EndFor
\For{$i$ \textbf{in} $range(len(H))$}
\If{$rest[i] == 0$}
\State $H'.append(H[i])$ 
\EndIf
\EndFor
\State \Return{$H'$} 
\EndFunction
\end{algorithmic} 
\label{alg:merge}
\end{algorithm}

\begin{algorithm}[H]
 \begin{algorithmic}[1] 
 \Require 
 Tree set: $H$, similarity matrix: $\widehat{{\mathbf S}}$.
 \Ensure Updated similarity matrix $\widehat{{\mathbf S}}$.
 \Function{Similarity}{$H, \widehat{{\mathbf S}}$}
 \State $n \gets len(H)$
 \State $\widehat{{\mathbf S}}' \gets n \times n$ empty matrix
 \For{$i$ \textbf{in} $range(n)$}
 \State $A \gets$ children index group of $H[i]$
 \For{$j$ \textbf{in} $range(n)$}
 \State $B \gets$ children index group of $H[j]$
 \State $\widehat{{\mathbf S}}'_{ij} \gets avg(\{\widehat{{\mathbf S}}_{ab}|a \in A, b \in B\})$
 \EndFor
 \EndFor
 \State \Return{$\widehat{{\mathbf S}}'$} 
 \EndFunction
 \end{algorithmic} 
\label{alg:similarity}
 \end{algorithm}
\end{minipage}
\caption{Computing Class Hierarchies}
\label{alg:CH}
\end{figure}

\section{Some Properties}

The following simple property says that when the given classification model has the
same similarity score for all pairs of base classes, then the computed hierarchy
simply groups all base classes into a superclass as the classifier is equally
confused about each pairs of them.

\begin{property}
If all the non-diagonal elements of $\mathbf S$ are the same, then
for any $r$ and $flag$, $CH({\mathbf S},r,flag)$ returns a tree that
has all the base classes as the children of the root.
\end{property}

To state the next property, we introduce a notion of {\em islands}.
Given a 
similarity matrix $\mathbf S$ for a set $C=\{0,...,{n-1}\}$ of classes, 
we say that a proper subset $X$ of $C$ is an island (under
$\mathbf S$) if
for any $i$ in $X$ and $j$ in $C\setminus X$, ${\mathbf S}_{ij} = 0$, and for any $i$ in X,
there is another $j$ in $X$ such that $i\neq j$ and ${\mathbf S}_{ij} \neq 0$.

It is easy to see that for a given similarity matrix, there may not exist an island.
Furthermore, if $X$ and $Y$ are two distinct islands, then $X\cap Y=\emptyset$.
This means that either $C$ has no island or there is a partition $\{ X_1,...,X_k\}$, $k>1$, of
$C$ such that each $X_i$ is an island of $C$.

\begin{property}
If $C$ has an island under ${\mathbf S}$, then there is a partition $\{ X_1,...,X_k\}$, $k>1$, of
$C$ such that each $X_i$ is an island of $C$, and 
for any $r$ and $flag$, $CH({\mathbf {S}},r,flag)$ returns a tree whose root has $k$
children $t_1,...,t_k$ such that the set of leaves in $t_i$ is $X_i$, $1\leq i\leq k$.
\end{property}

This property means that, for an island, its class hierarchy can be
constructed on its own independent of other classes, and when a set of classes is partitioned
into a collection of islands, the final hierarchy is just a superclass with each of
the island as a direct subclass.

Notice that as a special case, if a base class $i$ by itself is an island, then the given model
classifies it perfectly and our computed hierarchy will always put it directly under the root.

The basic idea behind our proof of this property is that suppose $X$ is an island,
then according to our algorithm, during
the construction of a hierarchy,  for any two trees $t$ and $t'$,
if the leaves in $t$ are all from $X$ and none of the leaves in $t'$ are from $X$, then
the similarity score between these two trees will be 0.

These two properties are useful to have but are by no means complete.
It's an interesting open theoretical question if
there is a complete set of properties that captures our algorithm in the sense that any
algorithm that satisfies the set of properties will be equivalent to ours.

\section{Experiments}

The key question that we had when starting this project was how good a class hierarchy can
one compute based on only a confusion matrix about a classifier. In the end, we were pleasantly
surprised that our algorithm came out with very informative ones. We describe in this
section the results of 
applying our algorithm in Figure~\ref{alg:CH} to the following classification tasks and models:
\begin{itemize}
\item Two well-known models ResNet10 \cite{he2016deep} and WideResNet28 \cite{wideresnet}
for CIFAR-10 \cite{cifar} image classification task.
\item Two models for two natural language classification tasks, one based on speech and the
other on texts.
\item A model for a music genre classification task based on music spectrogram.
\end{itemize}
In our experiment with these applications, we choose
the threshold ratio $r=0.1$ or $r=0.05$.
The smaller $r$ is, the more conservative our algorithm is when deciding when to
merge two classes.
To give an idea of how different values of $r$ affect the outcome, we include the results
of using $r=0,0.1,0.2,0.3,0.8,\geq 1$ on the WideResNet28 model for CIFAR-10 in the supplementary material.

\subsection{Image Classification}

\begin{figure}[htbp]
\subfigure[]
{
	\begin{minipage}{7cm}
	\centering        
	\includegraphics[width=3in]{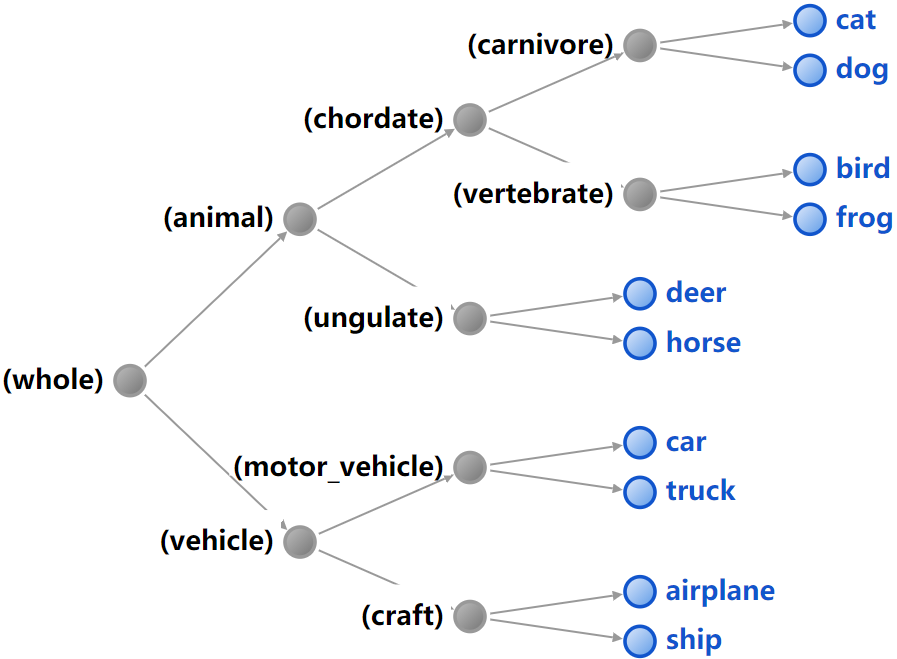}  
	\label{fig:cifar10_res10}
	\end{minipage}
}
\subfigure[]
{
	\begin{minipage}{7cm}
	\centering     
	\includegraphics[width=3in]{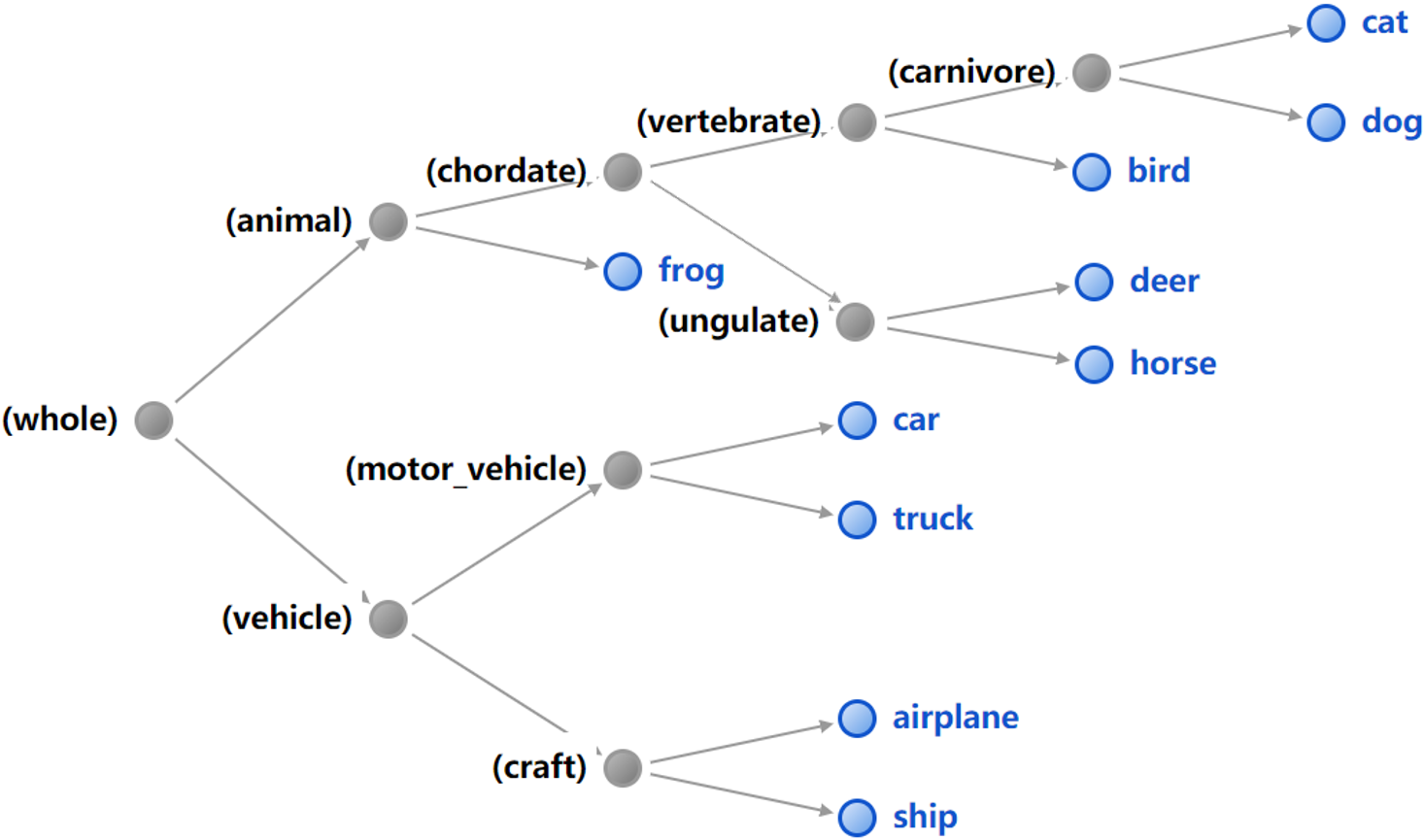}
	\label{fig:cifar10_wrn}
	\end{minipage}
}
\subfigure[]
{
	\begin{minipage}{7cm}
	\centering     
	\includegraphics[width=3in]{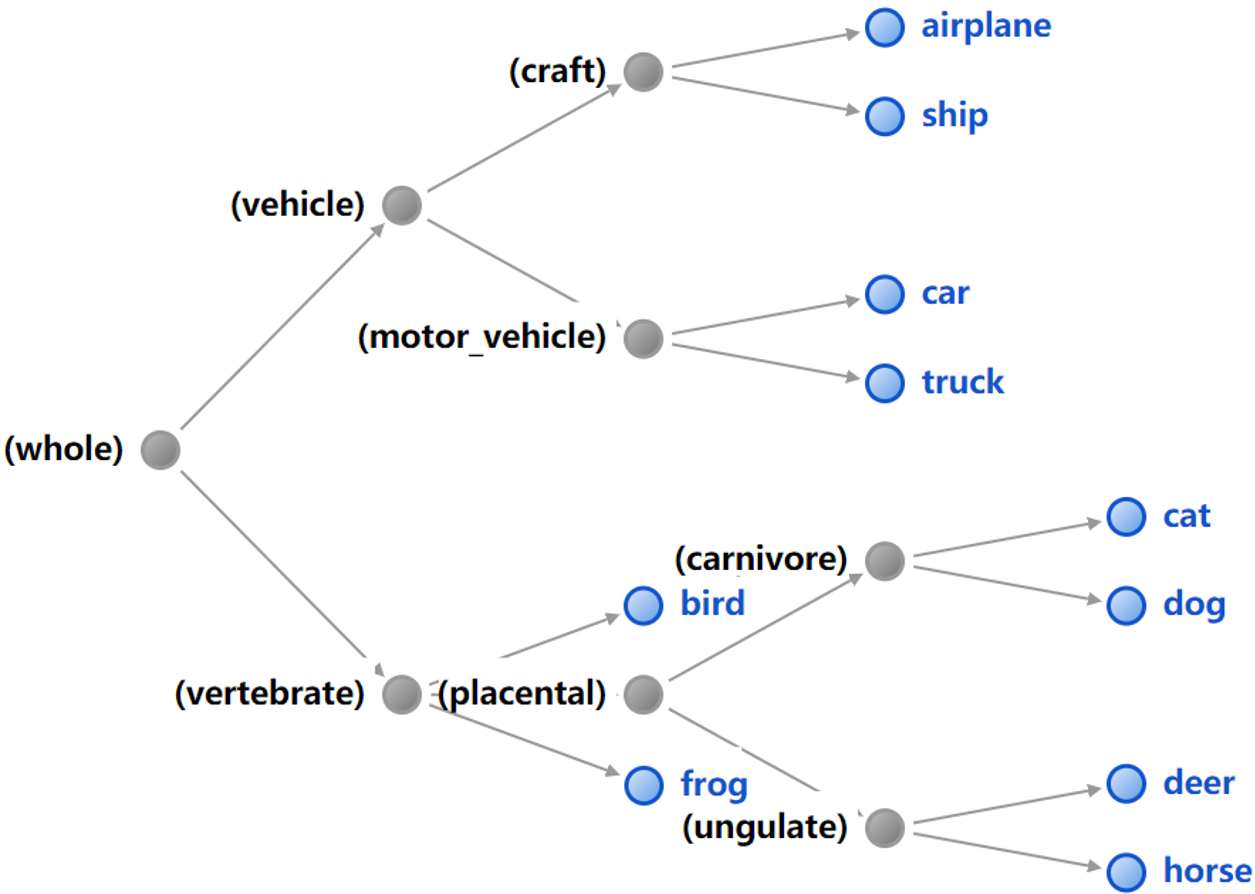}
	\label{fig:wordnet}
	\end{minipage}
}
\caption{The generated class hierarchy of ResNet10 (a) and WideResNet28 (b), and WordNet (c) hierarchy on CIFAR-10. For ease of comparison, the lables of internal nodes are in paranthesis
and obtained by searching the first common ancestor of each sub-tree's leaves from WordNet concept trees.}
\label{fig:resnet10-wide}
\end{figure}

CIFAR-10 is a widely used dataset for machine learning and computer vision. It contains 6000
images for each of the following 10 base classes:
airplanes, cars, birds, cats, deer, dogs, frogs, horses, ships, and trucks.
We considered two well-known CNNs, ResNet10 and WideResNet28, on the CIFAR-10 dataset.
Their accuracies are 93.64\% and 97.61\%, respectively. 
Figure \ref{fig:resnet10-wide} shows the SITs (single inheritance trees) returned by our
algorithm on them with $r=0.1$ ($flag=1$). 
As for the MITs for these two models, they have little differences with SITs in this task when $r=0.1$. More specifically, the MIT is the same with SIT for WideResNet28 model.

One can see that these two hierarchies are close, with major difference on the class ``frog''.
In the ResNet10 model, ``frog'' and ``bird'' have a confusion number high enough to be
combined into a superclass. However, under the WideResNet28 model, ``frog'' is well separated
from other ``chordates''.

Notice that the closeness of the SITs for the two models is despite the fact that one of them
has much higher accuracy than the other.
This shows the robustness of our algorithm, and also indicates that the resulting SITs are
indeed ``correct''.

Notice that ResNet10 and WideResNet28 are classification models based on images.
Since WordNet\footnote{\url{https://wordnet.princeton.edu}} includes conceptual-semantic
relations between English words, one can also use it to construct
a hierarchy of the 10 classes using their English labels.
The idea is that given a set of concepts in WordNet, one can find a most specific class that
subsumes these concepts. 
Figure \ref{fig:wordnet} is the hierarchy obtained this way for CIFAR-10 classes using WordNet.
We can see that all three hierarchies have the same structure for the
following 8 classes: airplanes, cars, cats, deer, dogs, horses, ships, and trucks.
For the remaining two, birds and frogs, the hierarchy computed from
WideResNet28 is closer to the one from WordNet.
It is certainly reassuring that the class hierarchies that are computed by our algorithm using
image-based classification models turn out to be similar with the concept hierarchy in a human
maintained database of conceptual-semantic relations.

Besides the pre-existing WordNet hierarchy, there has also been work on generating class
hierarchies by using some features inside a classification model. A recent work by
\citet{wan2020nbdt} proposes a method that
takes the weights of the last full-connection neural layer as the feature embeddings of
the classes, and use them to construct a class hierarchy, which they called it
a Neural-Backed Decision Tree (NBDT). They claimed that such NBDTs can improve
both the accuracy and interpretability of the original model.

\begin{figure}[htbp]
	\begin{minipage}{7cm}
	\centering     
	\includegraphics[width=3in]{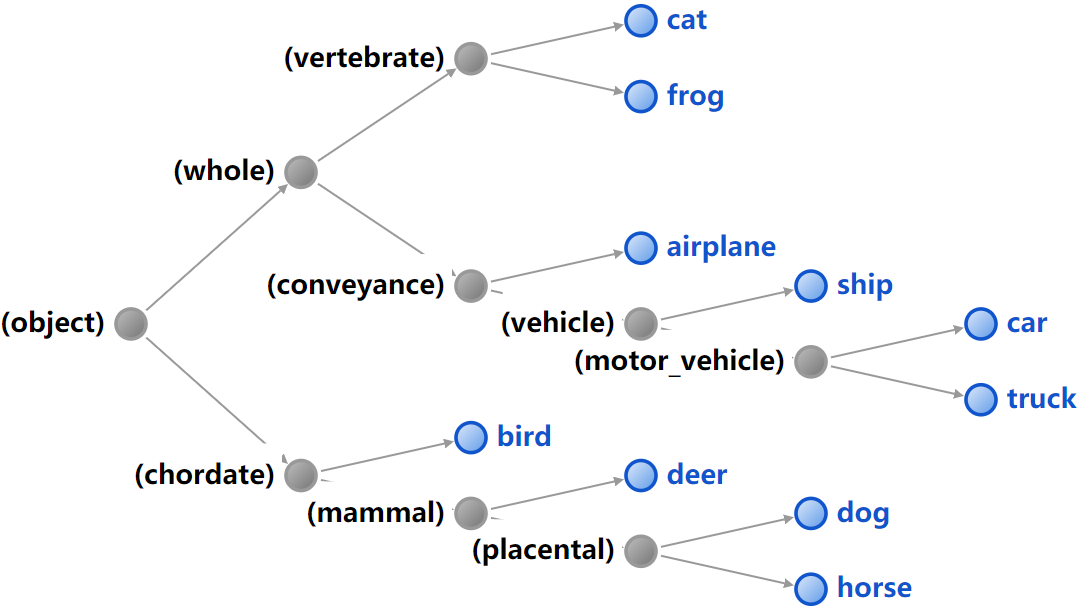}
	\end{minipage}
\caption{The NBDT induced by ResNet10.}
	\label{fig:nbdt-resnet10}
\end{figure}

Figure \ref{fig:nbdt-resnet10} is the NBDT from \cite{wan2020nbdt} using the ResNet10 model.
For the  WideResNet28 model, their NBDT turns out to be the same as
Figure \ref{fig:cifar10_res10}, the hierarchy computed
by our algorithm using the ResNet10 model.
So our algorithm generates the same hierarchy using a less accurate model.
Furthermore, considering that
our algorithm treats a model like ResNet10 as blackbox while the algorithm in
\cite{wan2020nbdt} uses weights inside a neural network, this result is definitely unexpected
and a pleasant surprise!

\citet{wan2020nbdt} 
proposed a new loss function based on their NBDTs to finetune the model and
showed that this sometimes can improve the accuracy of the original neural network.
In order to compare the hierarchies, we just replaced their NBDTs with our generated hierarchies (SITs) and finetuned the model to obtain our performance.
For CIFAR-10 and CIFAR-100, Table \ref{table2} shows the accuracies of ResNet10, ResNet18,
WideResNet28, as well as their revised models obtained by 
applying NBDTs and SITs.
Except for ResNet18 on CIFAR-100, using our generated hierarchies actually produce a slight boost.

\begin{table}[htbp]
\centering
\begin{tabular}{l|l|l|l}
	Method & Backbone & CIFAR-10 & CIFAR-100 \\
	\hline
	NN & resnet10 & 93.64 & 73.66 \\
	NN+SIT (ours) & resnet10 & \textbf{93.78} & \textbf{73.74} \\
	\hline
	NN & resnet18 & 94.74 & 75.92 \\
	NN+NBDT & resnet18 & 94.82 & \textbf{77.09} \\
	NN+SIT (ours) & resnet18 & \textbf{95.03} & 76.16 \\
	\hline
	NN & wrn28 & 97.62 & 82.09 \\
	NN+NBDT & wrn28 & 97.55 & 82.97 \\
	NN+SIT (ours) & wrn28 & \textbf{97.71} & \textbf{83.22} \\ 
\end{tabular}
\caption{The performance comparison of the original neural networks (NN), NN with their induced NBDTs (NN+NBDT) and NN with our generated hierarchis (NN+SIT). Wrn28 represents WideResNet28 model. \cite{wan2020nbdt} didn't provide their results on resnet10.}
\label{table2}
\end{table}

\subsection{Native Language Identification}
Our next experiment uses models for natural language classification. We consider two models,
one for identifying the speaker's native language from their English speech, and the other
for detecting the language given a text input.

\begin{figure}[htbp]
\centering
\includegraphics[width=0.45\textwidth]{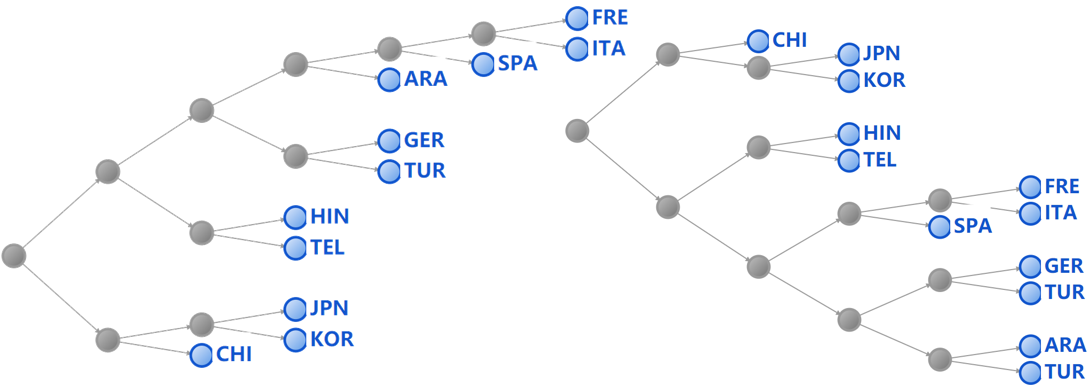} 
\caption{The generated class hierarchies for native language identification with speech as input with $r=0.05$ for both SIT (left) and MIT (right).}
\label{fig:langspoken}
\end{figure}

The first model is by \citet{spoken}. They built an LDA-based classifier
for detecting the speaker's native language from the recordings of their English speech.
The native languages considered are
Arabic (ARA), Chinese (CHI), French (FRE), German (GER), Hindi (HIN), Italian (ITA),
Japanese (JPN),
Korean (KOR), Spanish (SPA),  Telugu (TEL), and Turkish (TUR).
Figure \ref{fig:langspoken} shows the SIT and MIT generated from the confusion matrix of this model.
In both hierarchies,
Hindi and Telugu, which are two languages in India, are grouped together.
Chinese, Japanese, and Korean are in a separate branch from other languages.
The main difference between the SIT and MIT is that in the MIT, Turkish is duplicated once and
combined with Arabic, which seems to make more sense.

One possible use of the class hierarchies is in the teaching of English as a second language.
The hypothesis is that people whose native languages are closer according to the
hierarchy are likely to have some common issues when learning English and thus may be better
to be in the same class if necessary.

These experiments showed that our proposed algorithm can generate a very proper hierarchy
which could also be used to help to group non-native English speakers for the purpose of teaching English as a foreign language.

\begin{figure}[htbp]
\centering
\includegraphics[width=0.35\textwidth]{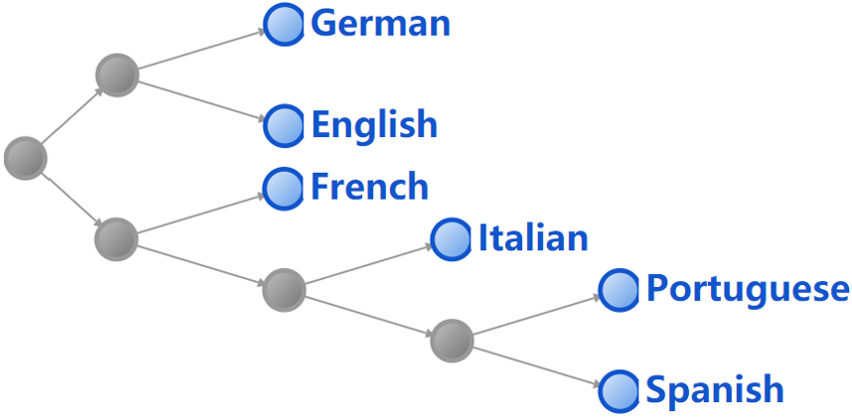} 
\caption{SIT for language identification task with text as input with $r = 0.05$.} 
\label{fig:langtext}
\end{figure}

The second model that we considered is the one by \citet{text}. They built a neural network
to detect the language of a text. The considered
six Latin languages: English, German, Spanish, French, Portuguese, and Italian.
Figure \ref{fig:langtext} shows the SIT computed by our algorithm using the confusion
matrix of the model with $r=0.05$.
We see the hierarchy groups German and English in one branch and the rest the other.
In fact, German and English belong to what is called Germanic languages \cite{enwiki:1042453262}
, and the rest Romance languages.
Again, it was a pleasant surprise to us that our fully automatic algorithm for building
a class hierarchy based only on the confusion matrix of a model somehow made the correct
classification. Of course, one can also say that this means the model by \citet{text} is
really good.

\subsection{Music Genre Classification}

\begin{figure}[htbp]
\centering
\includegraphics[width=0.4\textwidth]{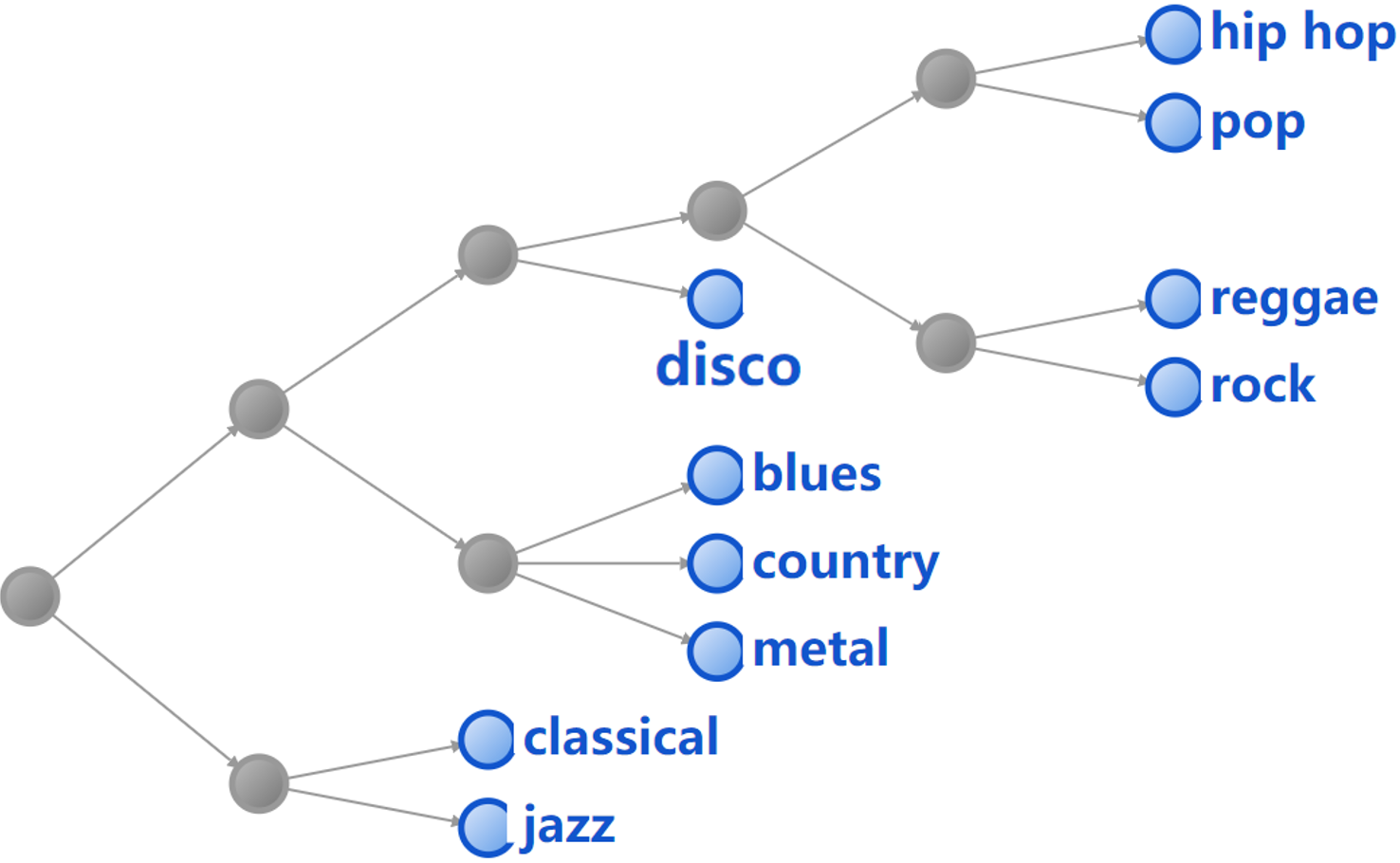} 
\caption{SIT for music genre classification task with $r = 0.05$.} 
\label{fig:music}
\end{figure}

Our next application is on music genre classification and the model is from
\citet{music}. It is a CNN model to classify the genre of the music based on its spectrogram.
Figure \ref{fig:music} is the hierarchy generated from the confusion matrix of this model.
It is interesting to see that it groups classical music and jazz together.
This means that in terms of spectrogram, classical music and jazz are more similar than
others. Indeed, according to the current Wikipedia jazz entry \cite{enwiki:1040542334},
``Jazz originated in the late-19th to early-20th century as interpretations of American and European classical music entwined with African and slave folk songs and the influences of West African culture''.

Assuming that people who like one type of music will likely like another type of music with
similar spectrogram, our generated hierarchy can be used for music recommendation.
For example, a music service can recommend blues to someone who listens to a lot of
country music, and vice versa.

\section{Related Works}

Class hierarchies are widely used to organize human
knowledge about concepts. So it is not surprising there has been much work
on utilizing this knowledge in AI. For instance, in machine learning, utilizing
word/concept hierarchies derived from the lexical database WordNet,
\citet{marszalek2007semantic} considered integrating prior knowledge about relationships among classes into the visual appearance learning. \citet{gao2011discriminative} applied a set of binary classifiers at each node of the hierarchy structure to achieve a good trade-off between accuracy and speed. \citet{brust2019integrating} integrates additional domain knowledge into classification and converted the properties of the hierarchy into a probabilistic model to improve existing classifiers.

Somewhat related is work on learning a decision tree from data \cite{Quinlan1983}. More
recently instead of raw data, 
\citet{wan2020nbdt} considered constructing a decision tree from
the weights of the last full-connection layer, and called it a
Neural-Backed Decision Tree (NBDT). We have compared their NBDTs with our hierarchies on the
two CIFAR-10 models. In terms of algorithms, they used the hierarchical agglomerative clustering algorithm from \cite{ward1963hierarchical}. This is an efficient clustering algorithm
that has been widely used in fields about document clustering \cite{zhao2005hierarchical}, multiple sequence alignment \cite{corpet1988multiple} and air pollution analysis \cite{govender2020application}. It constructs a binary tree starting from input data as leaves and merging
a pair of nodes step by step until only one node
is left, which represents the entire data set. Our algorithm shares this bottom up approach.
However, our algorithm is not restricted to binary trees, allows multiple inheritance, and
can be fine-tuned by the threshold ratio $r$.

\begin{figure}[htbp]
	\centering     
	\includegraphics[width=3in]{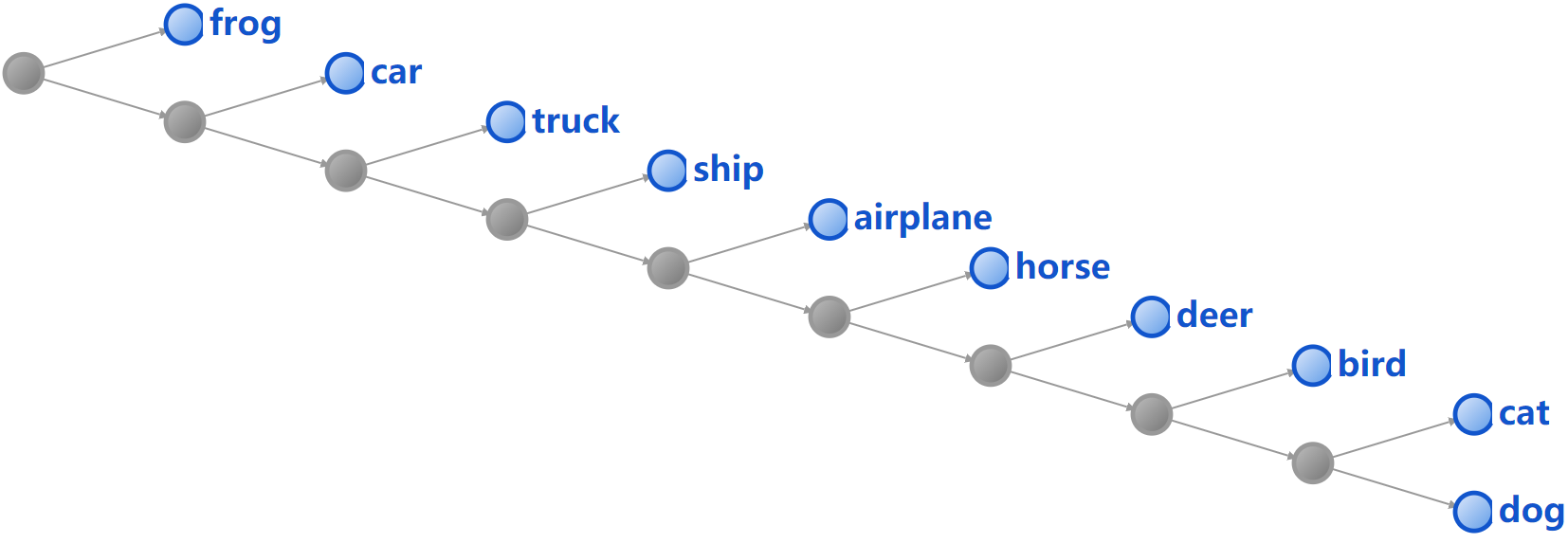}
\caption{The hierarchy generated using Euclidean Distance on the ResNet10 model for CIFAR-10}
	\label{fig:ed}
\end{figure}

Similar to our work, \citet{xiong2012building, cavalin2018confusion} also use
a model's confusion matrix to build a class hierarchy.
However, their algorithms are significantly different from ours.
\citet{cavalin2018confusion} used the following
Euclidean Distance (ED) to measure the distance between two classes:
\begin{center}
$\sqrt{\sum_{k=0}^{n-1} ({\mathbf M}_{ik} - {\mathbf M}_{jk})^2}$
\end{center}
Their idea is that if a model has  similar prediction distribution on $i$ and $j$, then the two classes should be very close. 
Also using prediction distribution, but instead of Euclidean distance,
\citet{xiong2012building} used the sum of absolute difference between two classes:
\[
\sum_{k=0}^{n-1} |{\mathbf M}_{ik} - {\mathbf M}_{jk}|.
\]
Instead of using a fixed matrix, our algorithm uses a dynamic criteria based on constraint
(\ref{eq:merge}).
As an example to illustrate the differences, Figure~\ref{fig:ed} shows the hierarchy generated
using the algorithm in \cite{cavalin2018confusion} on the ResNet10 model for CIFAR-10.
It is hard to justify such a hierarchy.

\section{Conclusion}

We have proposed an algorithm for computing a class hierarchy from a classification model.
Our algorithm treats the model as a black box, and uses only its so-called confusion matrix -
the matrix that records the number of errors that the model makes by misclassifying one class
as another. It generates surprisingly good hierarchies in a number of domains.
For CIFAR-10, it even generates the same hierarchy using ResNet10 as the neural-backed decision
tree of \cite{wan2020nbdt} using WideResNet28, despite that ResNet10 is not as accurate as
WideResNet28, and that we use only the confusion matrix while they used parameters inside
a neural network.

We believe this work provides an interesting new direction for connecting black box
classification models which are often large neural networks and symbolic
knowledge. For us, future work includes using computed class hierarchies {\em during} the
training process, as well as applying our algorithm for scientific discovery in some
large sets of classes for which there are still not good class hierarchies but enough data
to train some reasonably accurate classification models.
\bibliography{references}

\newpage
\section{Appendix}

This supplementary material includes two subsections. First is the detailed codes for functions we used in the $MERGE()$ function. The second subsection is to explore how important of parameter $r$ is to the generated hierarchies. 

\subsection{Algorithms Codes}
The detailed codes for function $FIND\_ALL\_PAIRS()$ are shown in figure \ref{alg:getpairslist}. It found all pairs satisfying condition
\begin{equation}
  {\mathbf S}_{ij} > 0 \textbf{ and } {\mathbf S}_{ij} + \delta \geq \max(\{{\mathbf S}_{i,k}| k \neq i\} \cup \{{\mathbf S}_{j,k}| k \neq j\})
\end{equation}
which we mentioned before and sorted them based on their similarity scores so that later the pair with higher score will be dealt earlier at every iteration.
\begin{figure}[hbtp]
\vspace*{-\baselineskip}
\begin{minipage}{\columnwidth}
\begin{algorithm}[H] 
\begin{algorithmic}[1] 
\Require 
$\widehat{{\mathbf S}}$ - similarity matrix.
$\delta$ - similarity threshold.
\Ensure 
$P$ - a list which includes the pair needs to be cared.
\Function{find\_all\_pairs}{$\widehat{{\mathbf S}}, \delta$}
\State $N \gets len(\widehat{{\mathbf S}})$
\State $P \gets []$
\State $sim \gets []$
\ForAll {$i,j$ such that $0 \leq i<j \leq N-1$}
\State $m' = \max(\{\widehat{{\mathbf S}}_{i,k}| k \neq i\} \cup \{\widehat{{\mathbf S}}_{j,k}| k \neq j\}$
\If{$\widehat{{\mathbf S}}_{ij} > 0$ \textbf{and} $\widehat{{\mathbf S}}_{ij} + \delta \geq m'$}
\State $P.append((i,j))$
\State $sim.append(\widehat{{\mathbf S}}_{ij})$
\EndIf
\EndFor
\State Sort $P$ based on the values in $sim$ in descending order
\State \Return{$P$} 
\EndFunction
\end{algorithmic} 
\end{algorithm}
\end{minipage}
\caption{Function: $FIND\_ALL\_PAIRS()$. Get list $P$ based on current similarity matrix.}
\label{alg:getpairslist}
\end{figure}

Next function $PAIRS\_TO\_GRAPHS()$ shown in figure \ref{alg:pairstographs} treated each pair as an edge of two nodes (superclasses in $H$), then merged them into bigger graph if and only if all the superclasses have pairs with other superclasses in this graph. Parameter $flag$ was used here to indicate the type of output tree (SIT or MIT). If $flag = 1$, then this function will return graphs without overlapping. Otherwise, the returned graphs may have overlapping. It is worth noting that, because of allowing overlapping when $flag = 0$, there may be a situation that graph $g_1$ is a subset of another graph $g_2$, in this case, only the bigger graph $g_2$ will be saved. Several other functions were used in this function. Their codes are shown in figure \ref{alg:functions}. 

\subsection{Similarity Threshold}
\begin{figure}[hbp]
\centering     
\includegraphics[width=3.3in]{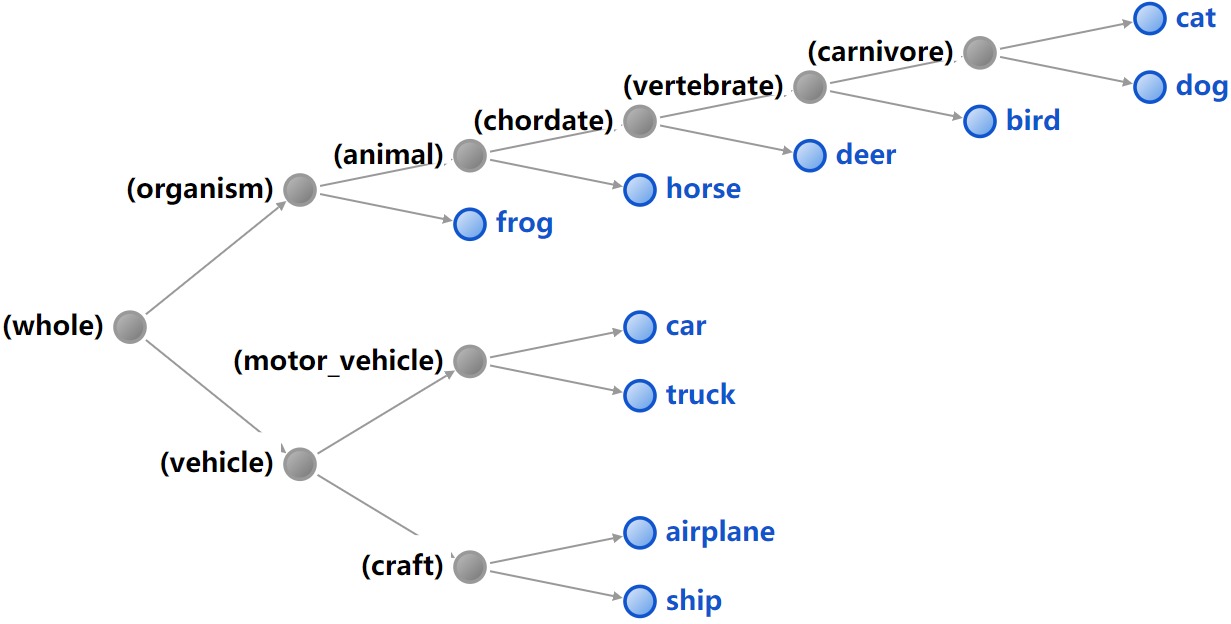}
\caption{The generated class hierarchy when $r =0$.}
\label{fig:cifar10_wrn_0}
\end{figure}

\begin{figure}[t]
\centering     
\includegraphics[width=3.3in]{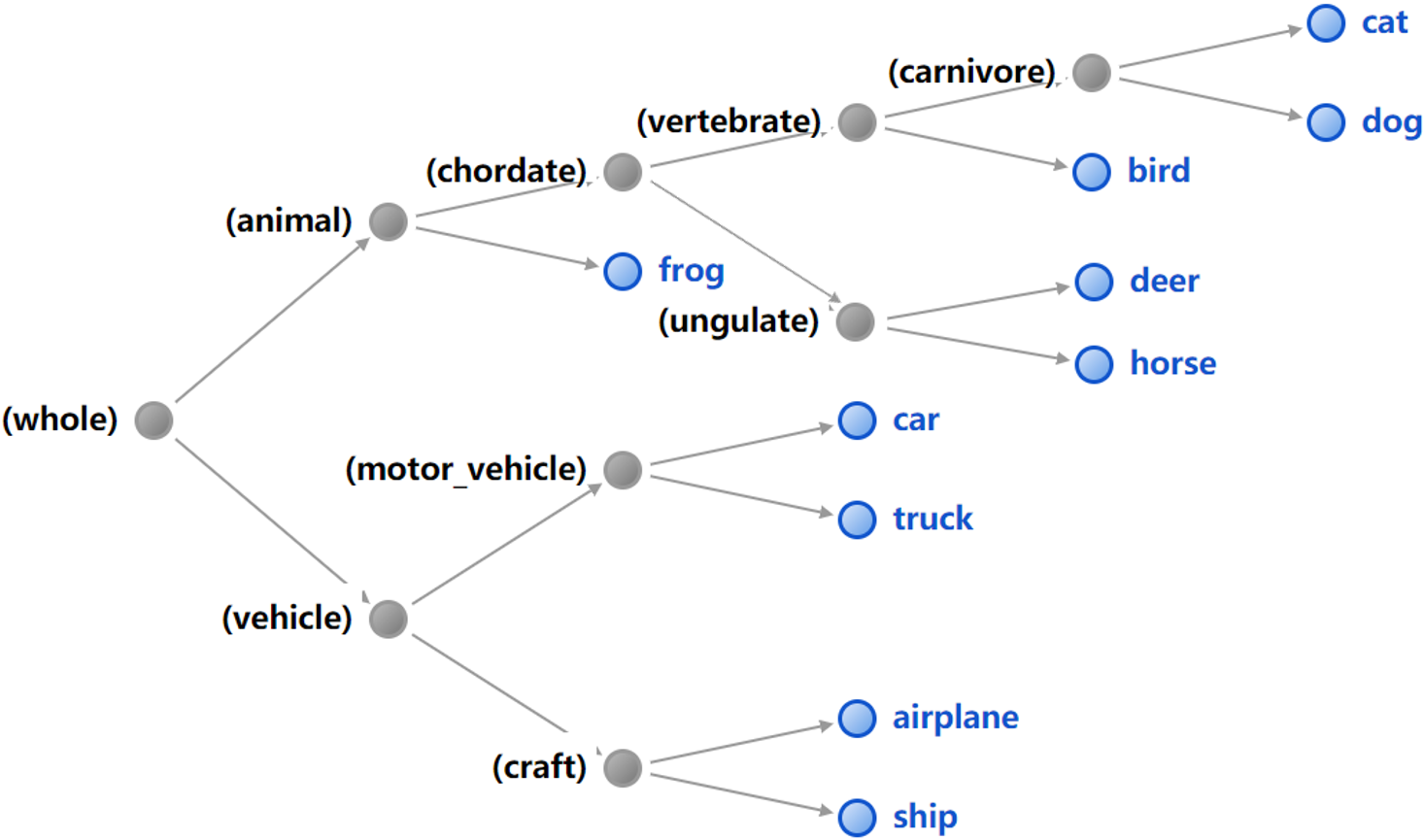}
\caption{The generated class hierarchy when $r = 0.1$.}
\label{fig:cifar10_wrn_1}
\end{figure}

\begin{figure}[htbp]
\centering     
\includegraphics[width=3.3in]{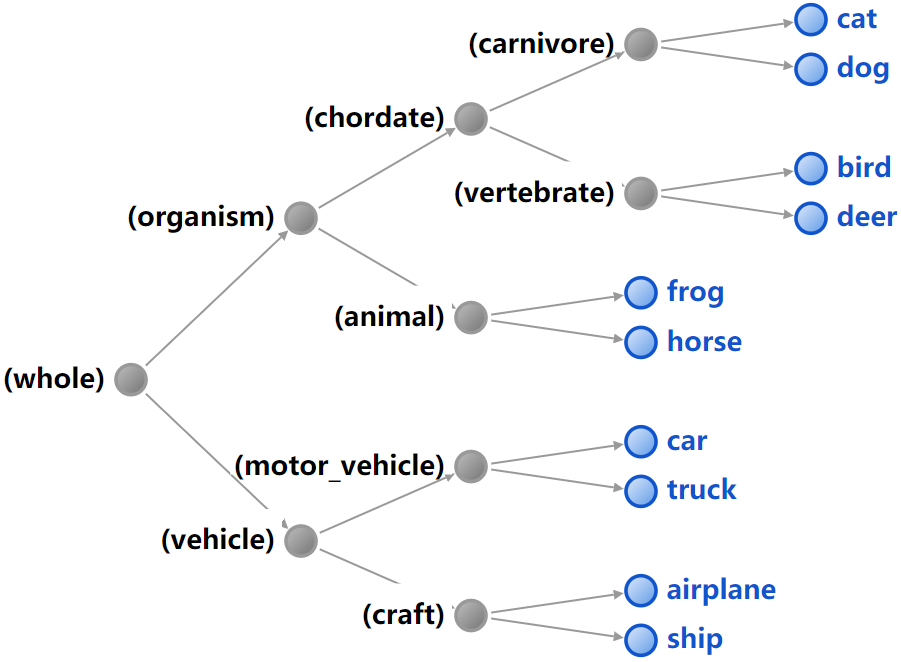}
\caption{The generated class hierarchy when $r = 0.2$.}
\label{fig:cifar10_wrn_2}
\end{figure}

\begin{figure}[htbp]
\centering     
\includegraphics[width=3.3in]{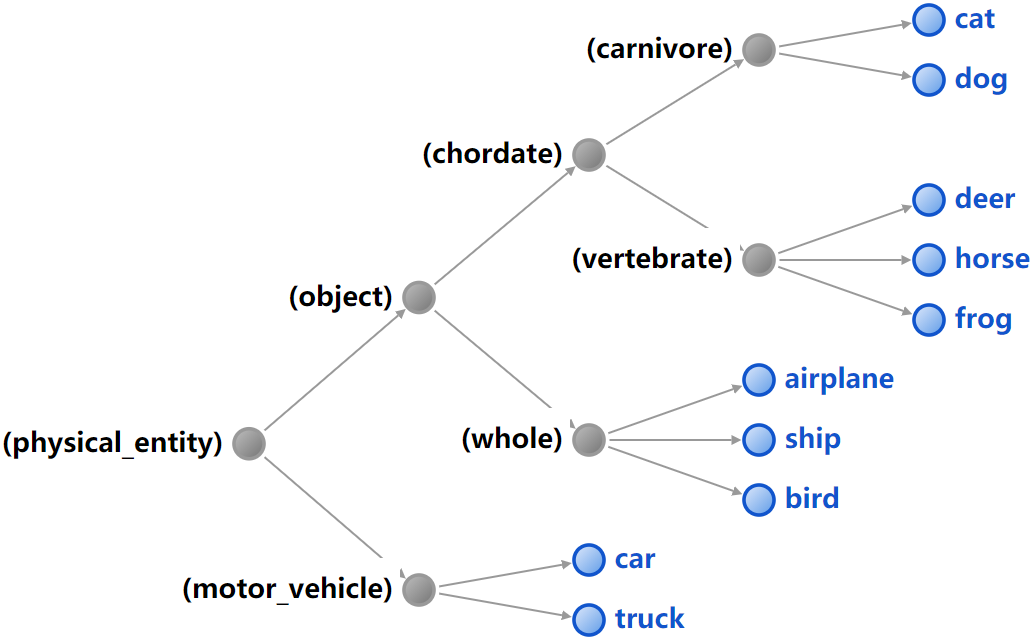}
\caption{The generated class hierarchy when $r = 0.3$.}
\label{fig:cifar10_wrn_3}
\end{figure}

\begin{figure}[htbp]
\centering     
\includegraphics[width=3.3in]{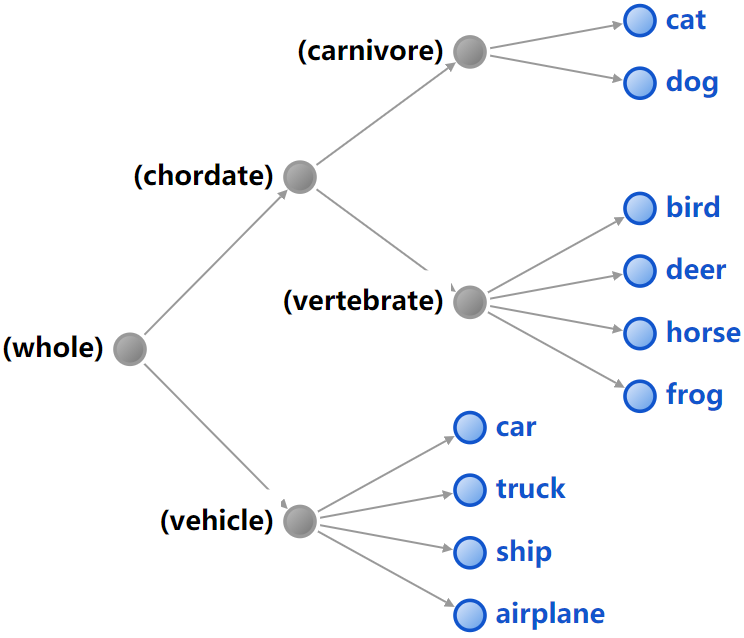}
\caption{The generated class hierarchy when $r = 0.8$.}
\label{fig:cifar10_wrn_8}
\end{figure}

\begin{figure}[htbp]
\centering     
\includegraphics[width=3.3in]{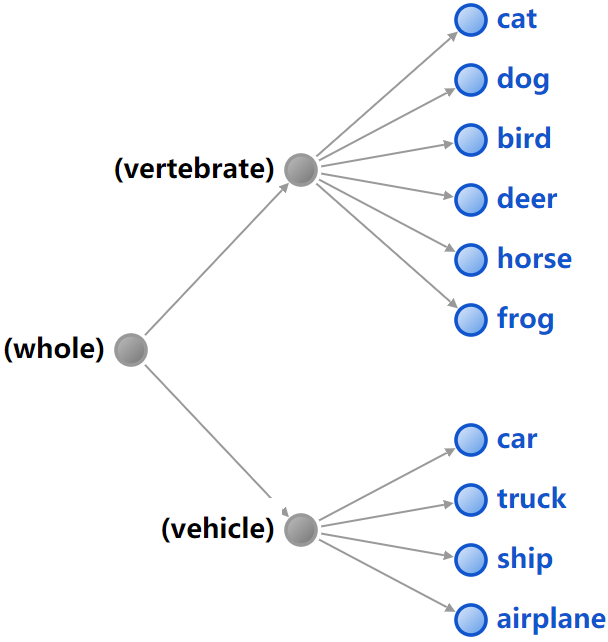}
\caption{The generated class hierarchy when $r \geq 1.0$.}
\label{fig:cifar10_wrn_10}
\end{figure}

We conducted experiments with WideResNet28 on CIFAR10 dataset to show different results of different $r$. Figure \ref{fig:cifar10_wrn_0} to \ref{fig:cifar10_wrn_10} have chosen several different $r$ and shown the hierarchies respectively. Notice that when $r \geq 1$, the generated tree will keep the same because all pairs with non-zero similarity score will be chosen at first iteration. From these results, we can see that the hierarchy tends to be shallower and wider with increasing $r$. Generally, a large $r$ means high tolerance for combination of classes, so it may cause some inaccurate combinations.
Consequently, a bigger $r$ will result a coarser tree with less hierarchical information. A low $r$ (e.g. 0.1) may be a good default value for most situations.

\begin{figure}[hbtp]
\vspace*{-\baselineskip}
\begin{minipage}{\columnwidth}
\begin{algorithm}[H] 
\begin{algorithmic}[1] 
\Require 
$\widehat{{\mathbf S}}$ - similarity matrix.
$P$ - pairs list. 
$flag$ - 1 if the output must be of single inheritance, 0 otherwise.
\Ensure 
$G$ - all fully connected graphs .
\Function{pairs\_to\_graphs}{$\widehat{{\mathbf S}}, P, flag$}
\State $G \gets []$
\State $n \gets len(\widehat{{\mathbf S}})$
\If{$flag$}
\While{$len(P) > 0$}
\State $D \gets \Call{get\_dict}{P, n}$
\State $g \gets P[0]$ 
\State $inter \gets \Call{get\_intersection}{g, D}$ 
\While{$len(inter) > 0$} 
\State $k \gets$ from $inter$ pick the one with highest similarity with $g$.
\State $g.append(k)$
\State $inter \gets \Call{get\_intersection}{g, D}$ 
\EndWhile
\State $P \gets P$ delete all pairs with index in $g$
\State $G.append(g)$
\EndWhile
\Else{}
\State $free \gets [True] * n$
\For{$i$ \textbf{in} $range(n)$}
\State $D \gets \Call{get\_dict}{P, n}$
\State $G' \gets \Call{find\_all\_graphs}{[i], [], D}$
\For{$g$ \textbf{in} $G'$}
\If{$len(g) == 1$}:
\If{$free[i]$}:
\State $G.append(g)$
\EndIf
\Else{}
\State $G.append(g)$
\EndIf
\For{$j$ \textbf{in} $g$}
\State $free[j] = False$
\EndFor
\EndFor
\State $P = [p$ \textbf{for} $p$ \textbf{in} $P$ \textbf{if} $i$ \textbf{not in} $p]$
\For{$j$ \textbf{in} $D[i]$}
\State $D[j].remove(i)$
\If{$set(D[j]).issubset(set(D[i]))$}
\State $P = [p$ \textbf{for} $p$ \textbf{in} $P$ \textbf{if} $j$ \textbf{not in} $p]$
\EndIf
\EndFor
\EndFor
\EndIf
\State \Return{$G$} 
\EndFunction
\end{algorithmic} 
\end{algorithm}
\end{minipage}
\caption{Function: $PAIRS\_TO\_GRAPHS()$. Find all fully connected graphs $G$ based on the $P$.}
\label{alg:pairstographs}
\end{figure}

\begin{figure}[hbtp]
\vspace*{-\baselineskip}
\begin{minipage}{\columnwidth}
\begin{algorithm}[H]
\begin{algorithmic}[1] 
\Require 
$P$ - pairs list. 
$n$ - current trees number.
\Ensure 
$D$ - pairs dictionary
\Function{get\_dict}{$P, n$}
\State $D \gets dict()$ 
\For{$i$ in $range(n)$}
\State $D[i] \gets []$
\EndFor
\For{$p$ in $P$}
\State $D[p[0]].append(p[1])$
\State $D[p[1]].append(p[0])$
\EndFor
\State \Return{$D$} 
\EndFunction
\end{algorithmic} 
\end{algorithm}

\begin{algorithm}[H] 
\begin{algorithmic}[1] 
\Require 
$g$ - base graph. 
$D$ - pairs dictionary. 
\Ensure 
$inter$ - all indexes of outside nodes which have edge with every node in $g$.
\Function{get\_intersection}{$g, D$}
\State $n \gets len(g)$
\State $inter \gets D[g[0]]$
\For{$i$ \textbf{in} range(1, $n$)}
\State $inter \gets list(set(inter) \& set(D[g[i]]))$
\EndFor
\State \Return{$inter$} 
\EndFunction
\end{algorithmic} 
\end{algorithm}

\begin{algorithm}[H] 
\begin{algorithmic}[1] 
\Require 
$g$ - base graph. 
$G$ - all graphs have been found. 
$D$ - pairs dictionary. 
\Ensure 
$G$ - all fully connected graphs with base graph $g$ .
\Function{find\_all\_graphs}{$g, G, D$}
\State $inter \gets \Call{get\_intersection}{g, D}$ 
\If{$len(inter) == 0$}
\State $g \gets sorted(g)$ 
\If{$g$ \textbf{not in} $G$}
\State $G.append(g)$ 
\EndIf
\Else
\For{$j$ \textbf{in} $inter$}
\State $G \gets \Call{find\_all\_graphs}{g + [j], G, D}$
\EndFor
\EndIf
\State \Return{$G$} 
\EndFunction
\end{algorithmic} 
\end{algorithm}

\end{minipage}
\caption{Several functions used in algorithm of figure \ref{alg:pairstographs}.}
\label{alg:functions}
\end{figure}

\end{document}